\documentclass[letter, 10 pt, conference]{ieeeconf}  

\IEEEoverridecommandlockouts % This command is only needed if 
\usepackage[english]{babel}
\usepackage[utf8]{inputenc}
\usepackage{times} % assumes new font selection scheme installed
\usepackage{cite}

\DeclareMathAlphabet{\pazocal}{OMS}{zplm}{m}{n}

\usepackage{hyperref}
\usepackage[nolist]{acronym}

\usepackage{multirow}

\usepackage{siunitx}
\usepackage{graphicx} % for pdf, bitmapped graphics files
\usepackage{epsfig} % for postscript graphics files
\usepackage{pgfplots}
\usepackage{caption}
\usepackage[font=footnotesize]{subcaption}
\usepackage{pgfplots}

\usepackage{float}

\usepackage[export]{adjustbox}

\usepackage{xcolor}
\definecolor{myYellow}{rgb}{0.93,0.69,0.13}
\definecolor{myPurple}{rgb}{0.49,0.18,0.56}
\definecolor{myGreen}{rgb}{0.26 0.72 0.54}

\usepackage{mathtools}
\usepackage{nicefrac}
\usepackage{amsmath}
\usepackage{amssymb}
\usepackage{bm} % to bold symbols

\DeclareMathOperator*{\minimize}{minimize}
\DeclareMathOperator*{\maximize}{maximize}

\usepackage{etoolbox}
\makeatletter%
\AfterPreamble{%
	\usepackage{hyperref}%
	\let\oldhypertarget\hypertarget%
	\renewcommand{\hypertarget}[2]{%
		\oldhypertarget{#1}{#2}%
		\protected@write\@mainaux{}{%
			\string\expandafter\string\gdef%
			\string\csname\string\detokenize{#1}\string\endcsname{#2}%
		}%
	}%
	\newcommand{\myhyperlink}[1]{%
		\hyperlink{#1}{\csname #1\endcsname}%
	}%
}
\makeatother%

\newcounter{Remark}
\newcounter{Definition}
\newcounter{Problem}
\usepackage{algorithm}
\usepackage[noend]{algpseudocode}

\makeatletter
\def\BState{\State\hskip-\ALG@thistlm}
\makeatother

\usepackage{tikz}
\pgfdeclarelayer{foreground}
\pgfsetlayers{background, main, foreground}
\usetikzlibrary{quotes, angles, backgrounds, arrows, automata, shapes, positioning, calc, through, spy, decorations.pathreplacing, decorations.markings, arrows.meta, automata, petri}

\tikzset{
    imglabel/.style={
      rectangle,
      inner sep=2pt,
      text=black,
      minimum height=1em,
      text centered,
      fill=white,
      fill opacity=1.0,
      text opacity=1,
      anchor=south west,
    },
  }
\tikzset{
	state/.style={
		rectangle,
		draw=black, very thick,
		minimum height=1.0em,
		text centered,
	},
}
\tikzset{
  on each segment/.style={
    decorate,
    decoration={
      show path construction,
      moveto code={},
      lineto code={
        \path [#1]
        (\tikzinputsegmentfirst) -- (\tikzinputsegmentlast);
      },
      curveto code={
        \path [#1] (\tikzinputsegmentfirst)
        .. controls
        (\tikzinputsegmentsupporta) and (\tikzinputsegmentsupportb)
        ..
        (\tikzinputsegmentlast);
      },
      closepath code={
        \path [#1]
        (\tikzinputsegmentfirst) -- (\tikzinputsegmentlast);
      },
    },
  },
  mid arrow/.style={postaction={decorate,decoration={
        markings,
        mark=at position .5 with {\arrow[#1]{stealth}}
      }}},
}
\graphicspath{{./figures/}}

\newcommand\copyrighttext{%
    \small \begin{center} \color{red} \textcopyright\,Accepted for presentation to the ``Formal methods techniques in robotics systems: Design and control" Workshop at IROS'23, Detroit, Michigan, USA. Personal use of this material is permitted. Permission from authors must be obtained for all other uses, in any current or future media, including reprinting/republishing this material for advertising or promotional purposes, creating new collective works, for resale or redistribution to servers or lists, or reuse of any copyrighted component of this work in other works. \end{center}}
\newcommand\copyrightnotice{%
	\begin{tikzpicture}[remember picture,overlay]
	\node[anchor=south,yshift=25.6cm] at (current page.south) 
	{\color{red}\fbox{\parbox{\dimexpr\textwidth-\fboxsep-\fboxrule\relax}{\copyrighttext}}};
	\end{tikzpicture}%
}

\title{\copyrightnotice \LARGE \bf Automating Bird Diverter Installation through Multi-Aerial Robots and Signal Temporal Logic Specifications}
\author{Alvaro Caballero$^{1\dag}$ and Giuseppe Silano$^{2\dag}$   
    \thanks{$^1$Department of System Engineering and Automation, University of Seville, 41092 Seville, Spain (email: {\tt alvarocaballero@us.es}).} 
    \thanks{$^2$Department of Cybernetics, Czech Technical University in Prague, 12135 Prague, Czech Republic (email: {\tt silangiu@fel.cvut.cz}).}
    \thanks{$^\dag$These authors contributed equally to this work. This publication is part of the R+D+i project TED2021-131716B-C22, funded by MCIN/AEI/10.13039/501100011033 and by the EU NextGenerationEU/PRTR. This work was also supported by the EU’s H2020 AERIAL-CORE grant no. 871479.}
}

\begin{document}

\maketitle
\thispagestyle{empty} % plain to see the number of pages
\pagestyle{empty} % plain to see the number of pages

%%% START SECTION ==========================================================

\begin{acronym}
    \acro{CBF}[CBF]{Control Barrier Function}
    \acro{MILP}[MILP]{Mixed-Integer Linear Programming}
    \acro{MPC}[MPC]{Model Predictive Control}
    \acro{RRT}[RRT]{Rapidly-exploring Random Tree}
    \acro{STL}[STL]{Signal Temporal Logic}
    \acro{UAV}[UAV]{Unmanned Aerial Vehicle}
    \acro{VRP}[VRP]{Vehicle Routing Problem}
    \acro{wrt}[w.r.t.]{with respect to}
\end{acronym}

%%% END SECTION ============================================================

%%% START SECTION ==========================================================

\begin{abstract}

This paper tackles the task assignment and trajectory generation problem for bird diverter installation using a fleet of multi-rotors. The proposed motion planner considers payload capacity, recharging constraints, and utilizes~\ac{STL} specifications for encoding mission objectives and temporal requirements. An event-based replanning strategy is introduced to handle unexpected failures and ensure operational continuity. An energy minimization term is also employed to implicitly save multi-rotor flight time during installation. Simulations in MATLAB and Gazebo, as well as field experiments, demonstrate the effectiveness and validity of the approach in a mock-up scenario.

% This paper tackles the task assignment and trajectory generation problem for bird diverter installation using a fleet of multi-rotors. The proposed motion planner considers payload capacity, recharging constraints, and utilizes~\ac{STL} specifications for encoding mission objectives and temporal requirements. To handle unexpected failures and ensure operational continuity, an event-based replanning strategy is introduced. Additionally, an energy minimization term is employed to implicitly save multi-rotor flight time during installation. Simulations in MATLAB and Gazebo, as well as field experiments demonstrate the effectiveness and validity of the approach in a mock-up scenario.

\end{abstract}

%%% END SECTION ============================================================

%%% START SECTION ==========================================================

\section{Full-version}
\label{sec:fullVersion}

% This avoids text going beyond the margins
\begin{sloppypar}
A full version of this work is available at \url{https://ieeexplore.ieee.org/document/10197369}. To reference, use~\cite{CaballeroAccess2023}.
% This avoids text going beyond the margins
\end{sloppypar}

%%% END SECTION ============================================================

%%% START SECTION ==========================================================

\section{Introduction}
\label{sec:introduction}

Power lines are critical infrastructure for supplying energy to millions of people. To enhance network reliability and reduce power outages, installation of bird diverters is crucial to mitigate the risk of bird collisions and improve visibility. %Bird mortality caused by power line collisions is a significant concern, impacting numerous bird species and resulting in millions of bird deaths annually worldwide. 
% Various bird diverter designs, including active and passive types, have been developed, along with visual and auditory deterrents. 
However, the current method of using manned helicopters for installation is time-consuming and poses safety risks~\cite{Ferrer2020}.

\acp{UAV} offer a promising solution for automating and replacing helicopters within the process.~\acp{UAV} can operate continuously over long distances and can be equipped with lightweight manipulation devices for autonomous operations~\cite{Suarez2021Access}. However, the limited battery and payload capacity of individual~\acp{UAV} require the use of multi-\ac{UAV} teams to expedite the process and cover large-scale scenarios. Planning for a multi-\ac{UAV} team presents challenges, including scheduling battery recharging, ensuring collision-free trajectories, and considering vehicle dynamics and energy consumption models.

Advanced task and motion planning techniques are necessary to enable bird diverter installation using multi-\ac{UAV} teams while meeting safety requirements and mission objectives.~\acf{STL}, a mathematical framework combining natural language commands with temporal and Boolean operators, can serve this purpose.~\ac{STL} is equipped with a \textit{robustness} metric, quantifying the extent to which system execution meets requirements~\cite{donze2010ICFMATS}. % Maximizing the robustness score leads to an optimization problem that provides the best feasible trajectory while satisfying desired specifications.

Therefore, this paper introduces a novel approach to task and motion planning for installing bird diverters on power lines using a team of multi-rotors. The proposed method leverages~\ac{STL} to generate optimal trajectories that satisfy mission requirements, considering vehicle dynamics, payload capacity limits, and installation time constraints. %The paper extends our previous research~\cite{Silano2021RAL} by incorporating payload capacity limitations and introducing an event-based replanning strategy to handle unforeseen failures. 
To ensure continuous operation, an event-based replanning strategy is introduced to handle unforeseen failures. Additionally, an energy minimization term is integrated to save multi-rotor flight time during installation operations. A hierarchical approach is adopt to handle the complexity of the resulting nonlinear optimization problem. First, a~\ac{MILP} problem is solved, and the resulting solution is fed into the final~\ac{STL} optimizer. % Numerical simulations in MATLAB, Gazebo simulations, and field experiments validate the method's effectiveness and feasibility. Videos that can be accessed at~\url{http://mrs.felk.cvut.cz/bird-diverter-ar}.

%%% END SECTION ============================================================

%%% START SECTION ==========================================================

\section{Problem Description}
\label{sec:problemDescription}

The installation of bird diverters involves visiting specific target regions along upper cables between consecutive towers. The~\acp{UAV} are assumed to be quadrotors with limited velocity, acceleration, and payload capacity. Ground-based refilling stations along the power line provide diverter reloading. The planning process considers vehicle dynamics, capacity constraints, obstacle avoidance, and safety requirements. The environment map, containing obstacles like power towers and cables, is assumed available before the mission.

%%% END SECTION ============================================================

%%% START SECTION ==========================================================

\section{Problem Solution}
\label{sec:problemSolution}

Le us define the state sequence $\mathbf{x}$ and the control input sequence $\mathbf{u}$ for the $d$-th multi-rotor as ${^d}\mathbf{x}=({^d}\mathbf{p}^{(1)}, {^d}\mathbf{v}^{(1)}, {^d}\mathbf{p}^{(2)}, {^d}\mathbf{v}^{(2)}, {^d}\mathbf{p}^{(3)}, {^d}\mathbf{v}^{(3)})^\top$ and ${^d}\mathbf{u}=({^d}\mathbf{a}^{(1)}, {^d}\mathbf{a}^{(2)}, {^d}\mathbf{a}^{(3)})^\top$, where ${^d}\mathbf{p}^{(j)}$, ${^d}\mathbf{v}^{(j)}$, and ${^d}\mathbf{a}^{(j)}$, with $j=\{1,2,3\}$, represent the sequences of position, velocity, and acceleration of the vehicle along the $j$-axis of the world frame, respectively. The $k$-th elements of ${^d}\mathbf{p}^{(j)}$, ${^d}\mathbf{v}^{(j)}$, ${^d}\mathbf{a}^{(j)}$, and $\mathbf{t}$ are denoted as ${^d}p^{(j)}_k$, ${^d}v^{(j)}_k$, ${^d}a^{(j)}_k$, and $t_k$, respectively, while the set of drones is denoted as $\mathcal{D}$. Hence, by employing the~\ac{STL} grammar (omitted here for brevity), the outlined bird installation problem is formulated with the~\ac{STL} formula:
\begin{equation}\label{eq:longRangeInspection}
    \resizebox{0.91\hsize}{!}{$%
    \begin{split}
    \varphi =&  \bigwedge_{d\in\mathcal{D}}\square_{[0,T_N]} ( {^d}\varphi_{\mathrm{ws}} \wedge {^d}\varphi_{\mathrm{obs}} \wedge {^d}\varphi_{\mathrm{dis}} ) \,
    \wedge \\
    & \bigwedge_{q=1}^\mathrm{tr}\lozenge_{[0,T_N-T_{\mathrm{ins}}]} \bigvee_{d\in\mathcal{D}}\square_{[0,T_{\mathrm{ins}}]} \left({^d}c(t_k) > 0 \right) {^d}\varphi_{\mathrm{tr,q}} \, \wedge \\
    & \bigwedge_{d\in\mathcal{D}} \lozenge_{[0,T_N-T_{\mathrm{rs}}]} \bigvee_{q=1}^\mathrm{rs}\square_{[0,T_{\mathrm{rs}}]}  \left({^d}c(t_k) = 0 \hspace{-0.3em} \implies \hspace{-0.3em} {^d}\mathbf{p}(t_k)\models {^d}\varphi_{\mathrm{rs,q}}\right) \wedge \\
    & \bigwedge_{d\in\mathcal{D}} \square_{[1,T_N-1]}\left({^d}\mathbf{p}(t_k)\models {^d}\varphi_{\mathrm{hm}} \hspace{-0.3em} \implies \hspace{-0.3em} {^d}\mathbf{p}(t_k+1)\models {^d}\varphi_{\mathrm{hm}} \right).
   \end{split}
    $}%
\end{equation}

The~\ac{STL} formula $\varphi$ comprises both safety and task requirements. The \textit{safety requirements} encompass three aspects: staying within the designated workspace (${^d}\varphi_\mathrm{ws}$), avoiding collisions with obstacles in the environment (${^d}\varphi_\mathrm{obs}$), and maintaining a safe distance from other~\acp{UAV} (${^d}\varphi_\mathrm{dis}$). On the other hand, the \textit{task requirements} focus on achieving specific tasks at predefined time intervals during the entire mission time $T_N$. Firstly, they guarantee that all target regions are visited by at least one~\ac{UAV} (${^d}\varphi_\mathrm{tr}$). Secondly, they ensure that each~\ac{UAV} remains in a target region for the designated installation time $T_\mathrm{ins}$, visits a refilling station, and stays there for a refilling time $T_\mathrm{rs}$ once they exhaust their onboard diverters (${^d}\varphi_\mathrm{rs}$). Finally, after completing their installation operations, each~\ac{UAV} should fly to the nearest refilling station (${^d}\varphi_\mathrm{hm}$). The following equations describe each of these specifications:
\begin{subequations}\label{eq:STLcomponents}
    \begin{align}
    \textstyle{{^d}\varphi_\mathrm{ws}} &= \textstyle{\bigwedge_{j=1}^3 \mathbf{p}^{(j)} \in (\underline{p}^{(j)}_\mathrm{ws}, \bar{p}^{(j)}_\mathrm{ws})}, \label{subeq:belongWorkspace} \\
    \textstyle{{^d}\varphi_\mathrm{obs}} &= \textstyle{ \bigwedge_{j=1}^3\bigwedge_{q=1}^{\mathrm{obs}} \mathbf{p}^{(j)} \hspace{-0.25em} \not\in (\underline{p}_{\mathrm{obs,q}}^{(j)}, \bar{p}_{\mathrm{obs,q}}^{(j)})}, \label{subeq:avoidObostacles} \\
    \textstyle{{^d}\varphi_\mathrm{dis}} &=  \textstyle{\bigwedge_{ \{d,m\} \in \mathcal{D}, d \neq m } \hspace{0.2em} \lVert {^d}\mathbf{p} - {^m}\mathbf{p} \rVert \geq \Gamma_\mathrm{dis}}, \label{subeq:keepDistance} \\
    \textstyle{{^d}\varphi_\mathrm{hm}} &= \textstyle{\bigwedge_{j=1}^3 \mathbf{p}^{(j)} \hspace{-0.25em} \in (\underline{p}^{(j)}_\mathrm{hm}, \bar{p}^{(j)}_\mathrm{hm})}, \label{subeq:backHome} \\
    \textstyle{{^d}\varphi_{\mathrm{tr,q}}} &=  \textstyle{\bigwedge_{j=1}^3  {^d}\mathbf{p}^{(j)} \hspace{-0.25em} \in (\underline{p}^{(j)}_{\mathrm{tr,q}}, \bar{p}^{(j)}_{\mathrm{tr,q}}) \, \wedge} \nonumber \\
    &  \textstyle{\wedge \bigcirc_{T_\mathrm{ins}} \left( {^d}c(t_k) = {^d}c(t_k-1) - 1 \right)}, \label{subeq:visitTargets} \\
    \textstyle{{^{d}}\varphi_\mathrm{rs,q}} &= \textstyle{\bigwedge_{j=1}^3 \mathbf{p}^{(j)} \hspace{-0.25em} \in (\underline{p}_\mathrm{rs,q}^{(j)}, \bar{p}_\mathrm{rs,q}^{(j)}) \, \wedge} \nonumber \\
    &  \textstyle{\wedge \bigcirc_{T_\mathrm{rs}} \left( {^d}c(t_k) = {^d}\bar{c} \right)}, \label{subeq:refill}
    \end{align}
\end{subequations} % \setcounter{equation}{3}
where $\underline{p}^{(j)}_\mathrm{ws}$, $\bar{p}^{(j)}_\mathrm{ws}$, $\underline{p}_\mathrm{obs,q}^{(j)}$, $\bar{p}_\mathrm{obs,q}^{(j)}$, $\underline{p}_\mathrm{hm}^{(j)}$, $\bar{p}_\mathrm{hm}^{(j)}$, $\underline{p}_\mathrm{tr,q}^{(j)}$, $\bar{p}_\mathrm{tr,q}^{(j)}$, $\underline{p}_\mathrm{rs,q}^{(j)}$, and $\bar{p}_\mathrm{rs,q}^{(j)}$  define the limits of the rectangular regions used for denoting workspace, obstacles, home, target and refilling stations areas; $\Gamma_\mathrm{dis} \in \mathbb{R}_{>0}$ represents the threshold value for the mutual distance between~\acp{UAV}; and ${^d}c(t_k)$ refers to the~\acp{UAV} payload capacity. The label $d$ is used to specify the particular drone to which the~\ac{STL} formula refers, while we refrain from using labels to indicate the vector stack of all drone variables.

Starting from mission specifications encoded as~\ac{STL} formula $\varphi$~\eqref{eq:longRangeInspection}, and replacing its robustness $\rho_\varphi(\mathbf{x})$ with the smooth approximation $\tilde{\rho}_\varphi(\mathbf{x})$, the generation of multi-rotor trajectories can be cast as an optimization problem~\cite{Silano2021RAL}:
%\footnote{To handle the computational complexity of non-differentiable functions like $\min$ and $\max$ in computing $\rho_\varphi(\mathbf{x})$, a beneficial approach is to use a smooth approximation, denoted as $\tilde{\rho}_\varphi(\mathbf{x})$, of the robustness function. This smoother approximation offers a more tractable and computationally efficient alternative.}
%
\begin{equation}\label{eq:optimizationProblemMotionPrimitives}
    \begin{split}
    &\maximize_{\mathbf{p}^{(j)}, \, \mathbf{v}^{(j)}, \, \mathbf{a}^{(j)} \atop d \in \mathcal{D}} \;\;
    {\tilde{\rho}_\varphi (\mathbf{p}^{(j)}, \mathbf{v}^{(j)} )} \\
    %
    %&\qquad \text{s.t.}~\quad \lvert {^d}v^{(j)}_k \rvert \leq {^d}\bar{v}^{(j)}, \lvert {^d}a^{(j)}_k \vert  \leq {^d}\bar{a}^{(j)}, \; \\
    %
    &\qquad \text{s.t.}~\quad\, {^d}\underline{v}^{(j)} \leq {^d}v^{(j)}_k \leq {^d}\bar{v}^{(j)},  \\
    &\,\;\;\, \qquad \quad\;\;\, {^d}\underline{a}^{(j)} \leq {^d}a^{(j)}_k \leq {^d}\bar{a}^{(j)}, \; \\
    &\,\;\;\, \qquad \quad\;\;\, \tilde{\rho}_\varphi (\mathbf{p}^{(j)}, \mathbf{v}^{(j)} ) \geq \varepsilon, \\
    &\,\;\;\, \qquad \quad\;\;\, {^d}\mathbf{S}^{(j)}, \forall k=\{0,1, \dots, N-1\},
    \end{split},
\end{equation}
where ${^d}\bar{v}^{(j)}$ and ${^d}\bar{a}^{(j)}$ represent the desired maximum values for velocity and acceleration, respectively, of drone $d$ along each $j$-axis of the world frame. The lower bound on robustness, $\tilde{\rho}_\varphi (\mathbf{p}^{(j)}, \mathbf{v}^{(j)}) \geq \varepsilon$, provides a safety margin for satisfying the~\ac{STL} formula $\varphi$ in the presence of disturbances. Finally, ${^d}\mathbf{S}^{(j)}$ refers to the motion primitives employed to describe the motion of drone $d$ along each $j$-axis~\cite{Silano2021RAL}.

However, the resulting problem is a nonlinear, non-convex max-min optimization problem and solving it within a reasonable time frame is challenging due to the propensity of solvers to converge to local optima~\cite{Bertsekas2012Book}. To address this issue, we compute an initial guess using a simplified~\ac{MILP} formulation on a subset of the original specifications $\varphi$:
\begin{subequations}\label{eq:MILP}
    \begin{align}
        % Cost function
        &\minimize_{z_{ij | d}, y_{j | d}}
        { \sum\limits_{ \{i,j\} \in \mathcal{V}, \, i \neq j, \, d \in \mathcal{D}} \hspace{-2em} w_{ij | d} \, z_{ij | d} } \label{subeq:objectiveFunction} \\
        %
        % Constraints (subject to part)
        &\;\;\;\;\;\; \text{s.t.} \;\, \hspace{-0.45cm} \sum\limits_{i \in \mathcal{V}, \, i \neq j, \, d \in \mathcal{D}} \hspace{-0.975em} z_{ij | d} = 2, \; \forall j \in \mathcal{T}, \label{subeq:visitedOnce} \\ % Codify that each node must be visited once
        &\quad \, \;\;\,\;\;\;\; \sum\limits_{ i \in \mathcal{V}, \, i \neq j} \quad \hspace{-0.975em} z_{ij | d} = 2y_{j | d}, \; \forall j \in \mathcal{T}, \; \forall d \in \mathcal{D}, \label{subeq:visitedOneUAV} \\ % Codify each node must be visited only by one UAV
        &\qquad \;\,\;\;\;\;\; \sum\limits_{ i \in \mathcal{T} } \qquad \hspace{-1.23em} z_{0i | d} = 1, \; \forall d \in \mathcal{D}, \label{subeq:depotVisitedOnce} \\  % The depot must be visited once by UAV (start mission)
        & \;\;\;\;\;\;\; \sum\limits_{\substack{ i \in \mathcal{T}, \, j \not\in \mathcal{T}, \, d \in \mathcal{D}}} \hspace{-1em} z_{ij | d} \geq 2 h\hspace{-0.2em}\left(\mathcal{T}\right). \label{subeq:capacityAndSubtours} % Capacity constraint (generalized subtour elimination constraint) %\; \forall \mathcal{S} \subset \mathcal{T},  
     \end{align}
\end{subequations}

The objective function~\eqref{subeq:objectiveFunction} quantifies the total distance covered by the team. Constraints~\eqref{subeq:visitedOnce} and~\eqref{subeq:visitedOneUAV} ensure that each target region is visited exactly once. To achieve this, auxiliary integer variables $y_{j | d} \in \{0,1\}$ are introduced, which ensure that if a~\ac{UAV} $d \in \mathcal{D}$ reaches target $j \in \mathcal{T}$, the same~\ac{UAV} must also leave the target. Constraint~\eqref{subeq:depotVisitedOnce} guarantees that each~\ac{UAV} starts the mission from its depot and does not return to it. Constraints~\eqref{subeq:capacityAndSubtours} serve two purposes: preventing tours that exceed the payload capacity of the~\acp{UAV} and ensuring that all tours connect to a refilling station, which is commonly known as the sub-tour elimination constraint. The lower bound $h(\mathcal{T})$ represents the minimum number of~\acp{UAV} required to visit all target regions $\mathcal{T}$.

As space is limited, we have excluded the mechanism for mission replanning in the event of~\ac{UAV} failures and the enhancement of the motion planner by minimizing energy consumption. For more information on these aspects, please refer to~\cite{CaballeroAccess2023}.

%To validate and assess the effectiveness of our proposed planning approach, we conducted a series of simulations and experiments. For more detailed information and visual demonstrations of the experimental results, we provide videos that can be accessed at~\url{http://mrs.felk.cvut.cz/bird-diverter-ar}.

%%% END SECTION ============================================================

%%% START SECTION ==========================================================

\section{Experimental Results}
\label{sec:experimentalResults}

The effectiveness and validity of the approach are demonstrated through simulations
in MATLAB and Gazebo, as well as field experiments carried out in a mock-up scenario. Videos that can be accessed at~\url{http://mrs.felk.cvut.cz/bird-diverter-ar}.

%%% END SECTION ============================================================

%%% START SECTION ==========================================================

\bibliographystyle{IEEEtran}
\bibliography{bib_short}

\end{document}